\documentclass[10pt,twocolumn,letterpaper]{article}
\usepackage[pagenumbers]{cvpr}

\usepackage{float}
\usepackage{xspace}
\usepackage{graphicx}
\usepackage{tabularx}
\usepackage{booktabs}
\usepackage{array}
\usepackage{makecell}
\usepackage{xcolor}
\usepackage{multirow}
\usepackage{mdframed}
\usepackage{tcolorbox}

\usepackage[hypcap=false]{caption}
\definecolor{customred}{RGB}{255, 99, 71}
\usepackage{tikz}
\usetikzlibrary{patterns}

\def\eg{\emph{e.g.}\xspace} 

\def\ie{\emph{i.e.}\xspace}

\newcommand\blfootnote[1]{%
\begingroup
\renewcommand\thefootnote{}\footnote{#1}%
\addtocounter{footnote}{-1}%
\endgroup
}

\definecolor{cvprblue}{rgb}{0.21,0.49,0.74}
\usepackage[pagebackref,breaklinks,colorlinks,allcolors=cvprblue]{hyperref}

\title{Pix2Cap-COCO: Advancing Visual Comprehension via Pixel-Level Captioning}

\author{
Zuyao~You$^{1,2*}$~\hspace{4pt}
Junke~Wang$^{1,2*}$~\hspace{4pt}
Lingyu~Kong$^{1}$~\hspace{4pt}
Bo~He$^{3}$~\hspace{4pt}
Zuxuan~Wu$^{1,2\dagger}$~\hspace{4pt}\\
$^{1}$Shanghai Key Lab of Intell. Info. Processing, School of CS, Fudan University \\
$^{2}$Shanghai Collaborative Innovation Center of Intelligent Visual Computing \\
$^{3}$University of Maryland, College Park
}

\begin{document}

\newcommand{\colorboxpattern}[1]{%
\begin{tikzpicture}
\fill[pattern=north east lines, pattern color=#1] (0,0) rectangle (0.6em,0.6em);
\draw[line width=0.5pt, color=#1] (0,0) rectangle (0.6em,0.6em);
\end{tikzpicture}%
}

\newcommand{\colordash}[1]{%
\begin{tikzpicture}
\draw[color=#1, line width=0.5pt, dash pattern=on 0.5pt off 0.5pt] (0,0) rectangle (0.6em,0.6em);
\end{tikzpicture}%
}

\twocolumn[{
\maketitle

\vspace{-1.4em}
\renewcommand\twocolumn[1][]{#1}
\begin{center}
\centering
\includegraphics[width=0.95\textwidth]{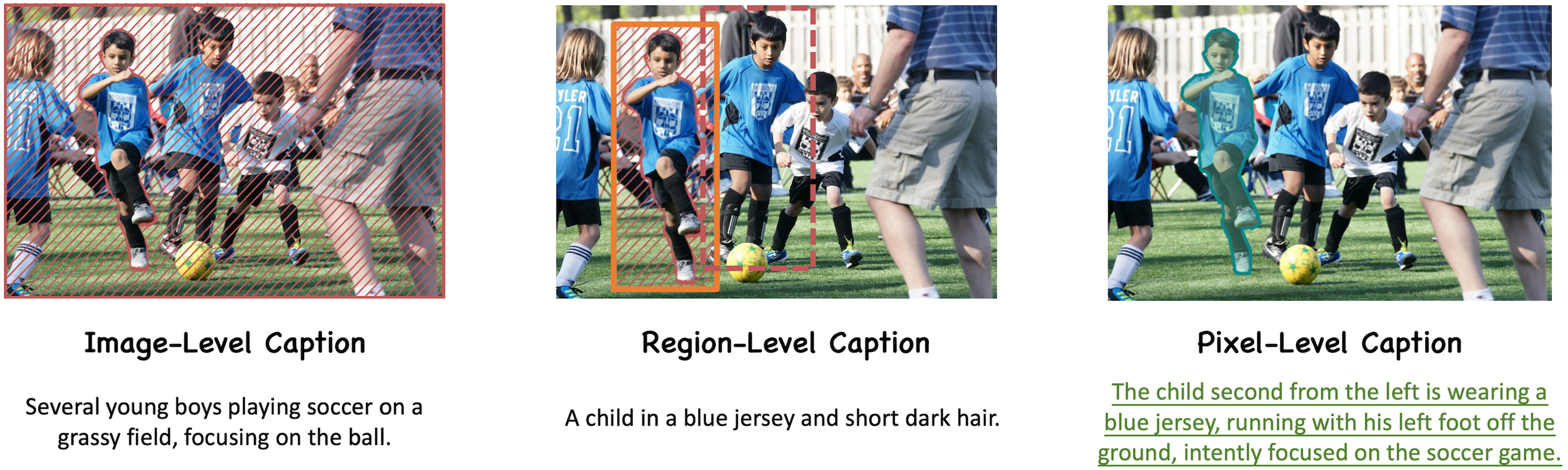}
\captionof{figure}{
Given a specified target (\eg, the second boy from the left), image-level caption and region-level caption both fail to align the visual input fully (\protect\colorboxpattern{customred} indicates the misalignment). Meanwhile, a one-sentence caption is sometimes too short to describe the target, which leads to confusion (\eg, the boy marked with \protect\colordash{customred}). In contrast, a pixel-level caption can provide precise, detailed information that aligns accurately with the visual input.
}
\label{tissue}
\end{center}
}
]

\blfootnote{$^{*}$ These authors contributed equally.}
\blfootnote{$^{\dagger}$ Corresponding author.}

\begin{abstract}
We present Pix2Cap-COCO, the first panoptic pixel-level caption dataset designed to advance fine-grained visual understanding. To achieve this, we carefully design an automated annotation pipeline that prompts GPT-4V to generate pixel-aligned, instance-specific captions for individual objects within images, enabling models to learn more granular relationships between objects and their contexts. This approach results in 167,254 detailed captions, with an average of 22.94 words per caption. Building on Pix2Cap-COCO, we introduce a novel task, panoptic segmentation-captioning, which challenges models to recognize instances in an image and provide detailed descriptions for each simultaneously. To benchmark this task, we design a robust baseline based on X-Decoder. The experimental results demonstrate that Pix2Cap-COCO is a particularly challenging dataset, as it requires models to excel in both fine-grained visual understanding and detailed language generation. Furthermore, we leverage Pix2Cap-COCO for Supervised Fine-Tuning (SFT) on large multimodal models (LMMs) to enhance their performance. For example, training with Pix2Cap-COCO significantly improves the performance of GPT4RoI, yielding gains in CIDEr (\(+1.4\%\)), ROUGE (\(+0.4\%\)), and SPICE (\(+0.5\%\)) on Visual Genome dataset, and strengthens its region understanding ability on the ViP-Bench, with an overall improvement of \(+5.1\%\), including notable increases in recognition accuracy (\(+11.2\%\)) and language generation quality (\(+22.2\%\)). Code is available at \href{https://github.com/geshang777/pix2cap}{\textcolor{magenta}{https://github.com/geshang777/pix2cap}}.
\end{abstract} 
\section{Introduction}
\label{sec:intro}
Existing datasets pairing visual inputs with descriptive text primarily focus on the image-level~\citep{young2014image,NIPS2011_5dd9db5e,sharma2018conceptual}, while effective for broader context, image-level descriptions lack grounding information that precisely ties objects to their locations within an image. Recent efforts have been made to develop region-level caption datasets~\citep{kazemzadeh2014referitgame,yu2016modeling,krishna2017visual}. However, these datasets fall short of fully aligning visual content with descriptions, as they always include irrelevant background information in the bounding boxes (see the middle of ~\cref{tissue}). Besides, the descriptions in these datasets are often too short of describing a specific object in the image, which may lead to confusion (\eg, the boy marked with \colordash{customred} in the second column of ~\cref{tissue}).

Creating a dataset with detailed pixel-level descriptions presents twofold challenges. Firstly, accurately recognizing and segmenting individual instances within an image requires precise delineation of object boundaries. Fortunately, this problem can be addressed effectively using existing high-quality object segmentation datasets, such as COCO~\citep{lin2014microsoft}, which provide robust annotations for object instances and their boundaries. These datasets serve as a strong foundation for training reliable segmentation models capable of handling diverse visual scenarios. Secondly, collecting high-quality, detailed, contextually precise captions that are capable of differentiating similar objects necessitates a deep understanding of the visual scene and nuanced language generation. While relying on human annotators could achieve high-quality results, the significant cost will undoubtedly pose a substantial barrier to scalability.

Recent years have witnessed the rapid development of large multimodal models (LMMs)~\citep{liu2024visual,wang2024qwen2,meng2024deepstack,li2024llava,wang2023chatvideo}. Among these, models like GPT-4V~\citep{openai2023gpt4v} have demonstrated exceptional capabilities in generating detailed and contextually rich textual descriptions based on visual inputs. Inspired by this, we propose an automated pipeline to annotate COCO~\citep{lin2014microsoft} with detailed pixel-level descriptions. Utilizing the masks provided by COCO~\citep{lin2014microsoft}, our approach begins by using the Set-of-Mark (SoM)~\citep{yang2023set} to mark and differentiate the objects within an image. These marked objects are paired with carefully designed prompts to guide GPT-4V in generating detailed captions for each instance. The resulting dataset, Pix2Cap-COCO, comprises 20,550 images and 167,254 captions with an average length of 8.14 words. We further divide it into a training set with 18,212 images, and a validation set with 2,338 images. Compared to existing region-level caption datasets~\citep{kazemzadeh2014referitgame,yu2016modeling,garg2024imageinwords,chung2024selective,krishna2017visual}, Pix2Cap-COCO stands out for offering significantly richer linguistic diversity and more precise pixel-level annotations, while also achieving a comparable scale. To improve the caption quality further, we recruited human annotators to manually refine and correct errors in key object attributes, such as color, to establish a highly reliable validation set.

With the proposed Pix2Cap-COCO dataset, we introduce a novel task called panoptic segmentation-captioning, where models are required to produce panoptic segmentation masks and generate pixel-level captions for each mask. Compared to panoptic segmentation which focuses on categorizing and segmenting visual elements~\citep{kirillov2019panoptic}, or image captioning that aims to generate descriptions for an entire image~\citep{chen2015microsoft,krishna2017visual}, panoptic segmentation-captioning presents a greater challenge by requiring the seamless integration of visual segmentation and language generation. We believe that it can facilitate the advancement of instance-level understanding~\citep{zou2023generalized,wang2023look,wang2023omnitracker,you2025focus} by empowering models to comprehensively analyze and describe visual scenes. To resolve this task, we design a baseline by extending X-Decoder~\citep{zou2023generalized} with an additional captioning head, which can learn the prediction of panoptic segmentation masks and pixel-level captions in an end-to-end manner.

To further evaluate the utility of Pix2Cap-COCO, we incorporate it to train GPT4RoI~\citep{zhang2023gpt4roi}, a large multimodal model that leverages region-based visual inputs to achieve detailed instance-level understanding, and observe significant performance improvements across multiple benchmarks. For example, using Pix2Cap-COCO improves GPT4RoI by 5.1\% on average on ViP-Bench~\citep{cai2024vipllava}, with notable gains in recognition accuracy (+11.2\%) and language generation quality (+22.2\%). These performance improvements underscore the significance of Pix2Cap-COCO as a high-quality source for fine-grained alignment between visual and textual representations.

\section{Related Work}

\paragraph{Datasets for Visual Understanding.}
Advancements in computer vision research fundamentally depend on the availability of large-scale, high-quality datasets. In the domain of visual captioning, early datasets such as COCO~\cite{lin2014microsoft}, Flickr30K~\citep{young2014image}, and VizWiz~\citep{gurari2018vizwiz} lay the groundwork for visual understanding by offering short, single-sentence descriptions of entire images. Building on this, the evolution of annotation granularity gives rise to region-level captioning datasets~\citep{krause2017hierarchical,chung2024selective,urbanek2024picture} designed for dense captioning tasks. Notably, Visual Genome~\citep{krishna2017visual} provides extensive object descriptions paired with bounding boxes, paving the way for instance captioning and fine-grained visual understanding. Subsequent efforts, such as the RefCOCO series (RefCOCO~\citep{kazemzadeh2014referitgame}, RefCOCO+\citep{yu2016modeling}, and RefCOCOg\citep{yu2016modeling}), emphasize natural language expressions for object localization to advance the intersection of vision and language. Despite these advancements, existing datasets are still limited to image-level or region-level, and lack the alignment between pixels and textual descriptions. To address this gap, our work introduces a novel pixel-level captioning dataset that establishes a precise correspondence between text and instance masks. This dataset provides a new benchmark for developing models capable of capturing intricate visual-textual relationships.

\begin{figure*}[t]
\centering
\includegraphics[width=2.0\columnwidth]{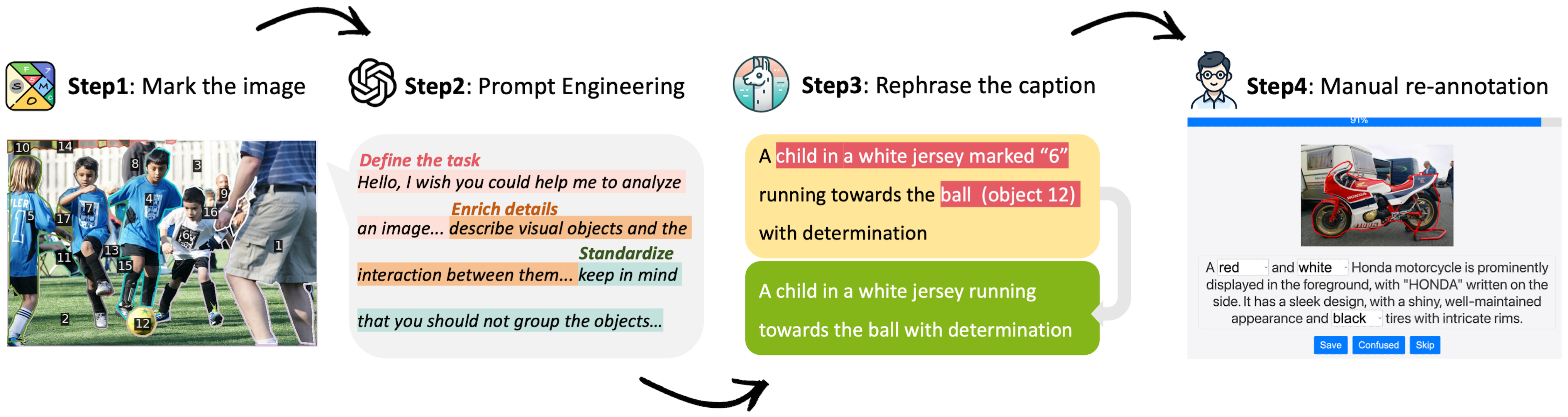}
\caption{Overview of the Pix2Cap-COCO dataset pipeline. \textbf{Step 1:} Apply COCO mask annotations to the image; \textbf{Step 2:} Engineer prompts to generate detailed, formatted pixel-level captions; \textbf{Step 3:} Refine captions through hard matching and rephrasing with LLaMA; \textbf{Step 4:} Conduct manual re-annotation to establish an accurate benchmark.}
\label{fig:fdatasetpipeline}
\end{figure*}

\paragraph{GPT-4V for Dataset Generation.}
Recent advancements in large multimodal models, particularly GPT-4V~\citep{openai2023gpt4v}, have unlocked new possibilities for automating dataset annotation, accelerating the creation of datasets that traditionally require extensive manual effort. LVIS-Instruct4V~\citep{wang2023see} and ShareGPT4V~\citep{chen2023sharegpt4v} prompt GPT-4V to automatically generate question-answer pairs for given images, aiming to improve the supervised fine-tuning data of LLaVA~\citep{liu2024visual}. ShareGPT4Video~\citep{chen2024sharegpt4video} extend this strategy to video datasets, to construct a high-quality dataset for video-based visual-language tasks. These advancements collectively highlight the transformative potential of GPT-4V in improving dataset creation workflows across visual and video domains. However, the above work focuses on high-level image or video captioning without pixel-level details, failing to meet the growing demand for finer-grained visual understanding in LLMs~\citep{rasheed2024glamm,zhang2024omg,ren2024pixellm}. Comparatively, our work presents the first panoptic pixel-level dense captioning dataset utilizing GPT-4V, advancing the visual comprehension ability of models to the next level.
\section{Pix2Cap-COCO}
\subsection{Annotation Pipeline}
COCO~\citep{lin2014microsoft} is a widely used dataset for object detection and segmentation, containing 1.5 million instances covering a wide range of object categories. In this work, we build our pixel-level caption dataset upon COCO by directly using the panoptic segmentation masks they provide as the pixel annotations. To obtain the pixel-aligned dense captions, we first mark the instances in an image with unfilled polygons, with a unique bright object ID at the center, following~\citep{yang2023set}. In this way, instances in each image are uniquely marked, enabling precise differentiation of different objects.

Subsequently, the images are processed with GPT-4V~\citep{openai2023gpt4v}, guided by a carefully designed text prompt to generate detailed and context-aware captions.

\begin{tcolorbox}[colframe=black, colback=gray!10, rounded corners, width=\linewidth]
\small
\textit{You are an AI visual assistant analyzing a single image. In the given image, I label \texttt{\(<\)num instances\(>\)} objects (or background) by marking each with a bright numeric ID at its center and boundary. I can also provide their categories: \texttt{\(<\)img categories\(>\)}.}
\end{tcolorbox}

\texttt{\(<\)num instances\(>\)} and \texttt{\(<\)img categories\(>\)} here are placeholders that are replaced by the number of instances and categories for each image. To ensure object-specific captions and avoid generic descriptions, we add:

\begin{tcolorbox}[colframe=black, colback=gray!10, rounded corners, width=\linewidth]
\small

\textit{I aim to use your descriptions for a grounding dataset, so each object description should be unique. For instances of the same category, provide detailed descriptions to differentiate them. Include appearance, interactions, and details such as color, shape, texture, emotion, motion, intention, object counts, position, and relative position.}
\end{tcolorbox}

To streamline caption generation for subsequent pipeline stages, we impose specific constraints:

\begin{tcolorbox}[colframe=black, colback=gray!10, rounded corners, width=\linewidth]
\small

\textit{Your answer should contain details and follow the following format:
The objects include: 
object id. summary of the object: description of the object (e.g., 1. Person: the person is wearing a pink jacket, black pants, and a pink beanie. They are holding ski poles and are skiing on a snowy mountain.)}
\end{tcolorbox}

\begin{table*}[t!]
\centering
\caption{As the first panoptic pixel-level captioning dataset, Pix2Cap-COCO has considerable data scale and caption quality. We compare with previous datasets on overall statistics and per caption analysis below.}
\label{table:linguistic}
\setlength{\tabcolsep}{1.0mm}
\renewcommand{\arraystretch}{1.1}
\footnotesize
\begin{tabular}{l|c|*{3}{c}|*{7}{c}}
\toprule
{} & \multicolumn{4}{c}{Overall Statistics} & \multicolumn{7}{c}{Per-Caption Linguistic Analysis } \\
\cmidrule[0.5pt](rl){2-5}
\cmidrule[0.5pt](rl){6-12}
& \textbf{Visual Input} & \textbf{Images}& \textbf{Captions}& \textbf{Cap/Img}& \textbf{Characters}& \textbf{Words}& \textbf{Sentences} & \textbf{Noun}  & \textbf{Adjective}  &  \textbf{Adverb}  &  \textbf{Verb}  \\
\midrule
\textbf{VisArgs~\citep{chung2024selective}} &bbox  &1,611 &5,112   &3.17  &\underline{79.14}  &\underline{14.11} &\underline{1.61} &\underline{4.89} &\underline{1.12} &\underline{0.16} &\underline{2.44} \\
\textbf{Visual Genome~\citep{krishna2017visual}} &bbox  &\textbf{108,077} &\textbf{5,408,689} &\textbf{50.04} &25.51 &5.09 &1.00    &2.24 &0.61 &0.03 &0.47 \\
\textbf{RefCOCO~\citep{kazemzadeh2014referitgame}}       &bbox  &19,994  &142,210   &7.11  &17.72 &3.50 &1.00    &1.76 &0.55 &0.13 &0.26 \\
\textbf{RefCOCO+~\citep{yu2016modeling}}      &bbox  &19,992  &141,564   &7.08  &18.41 &3.53 &1.00    &1.69 &0.61 &0.09 &0.26 \\
\textbf{RefCOCOg~\citep{yu2016modeling}}      &bbox  &\underline{25,799}  &95,010    &3.68  &41.31 &8.46 &1.06 &3.03 &0.99 &0.08 &0.74 \\
\midrule
\textbf{Ours}   &mask  &20,550 &\underline{167,254} &\underline{8.14} &\textbf{138.13} &\textbf{22.94} &\textbf{2.73}  &\textbf{7.08} &\textbf{3.46} &\textbf{0.54} &\textbf{3.42}     \\
\bottomrule
\end{tabular}
\end{table*}

\begin{figure*}[t!]
\centering
\includegraphics[width=0.95\linewidth]{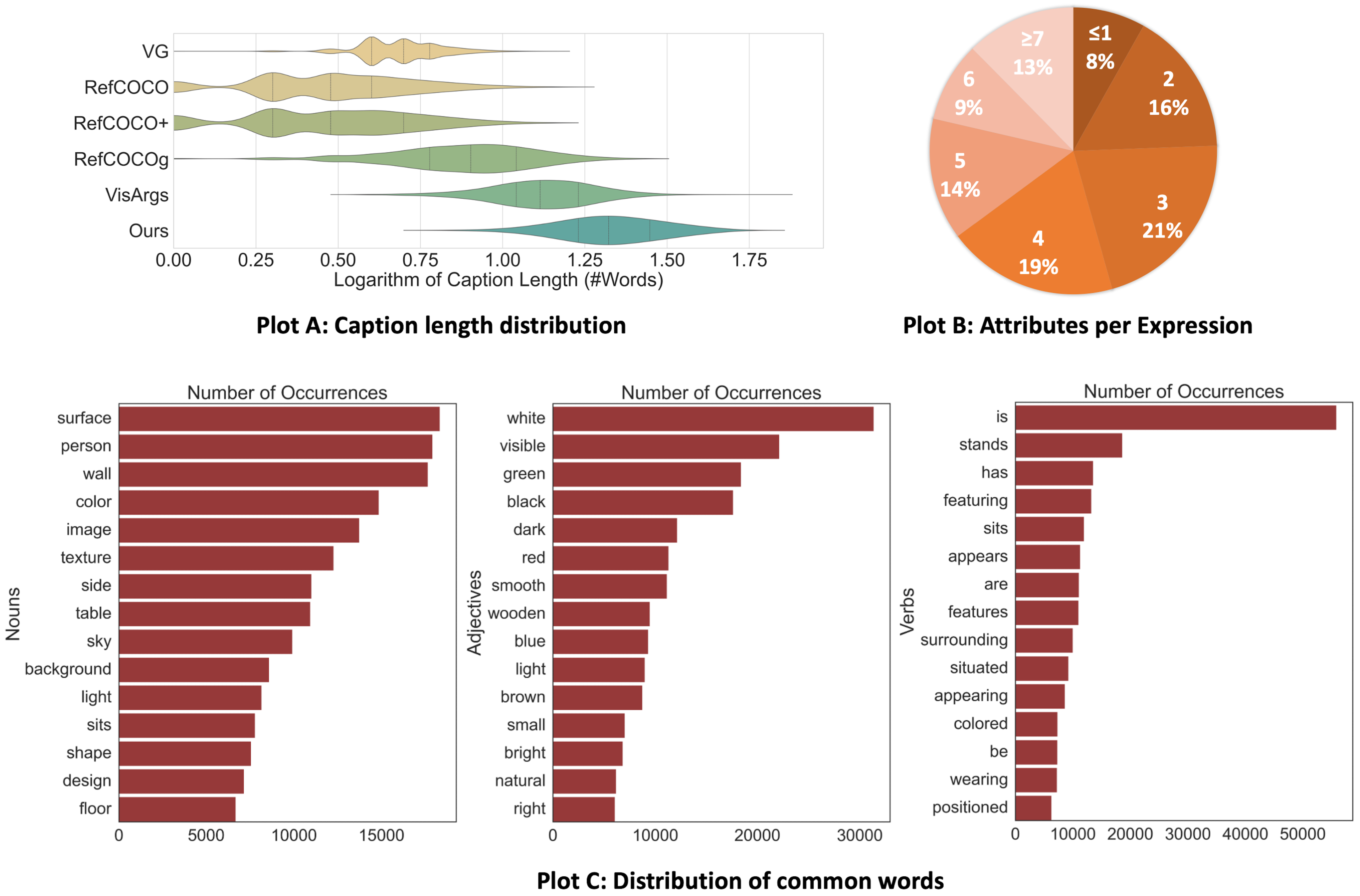}

\caption{Analyses on Pix2Cap-COCO. \textbf{Plot A} shows the caption length distribution, with vertical dotted lines in the violin plot representing quartiles. We use \(\log_{10}(\text{words per caption})\) as the horizontal axis to deal with the long-tail distribution of caption lengths. \textbf{Plot B} demonstrates the richness of attributes in our captions. \textbf{Plot C} is the distribution of the most common words in Pix2Cap-COCO captions.}
\label{fig:combine}
\end{figure*}

\begin{figure*}[t]
\centering
\includegraphics[width=2.0\columnwidth]{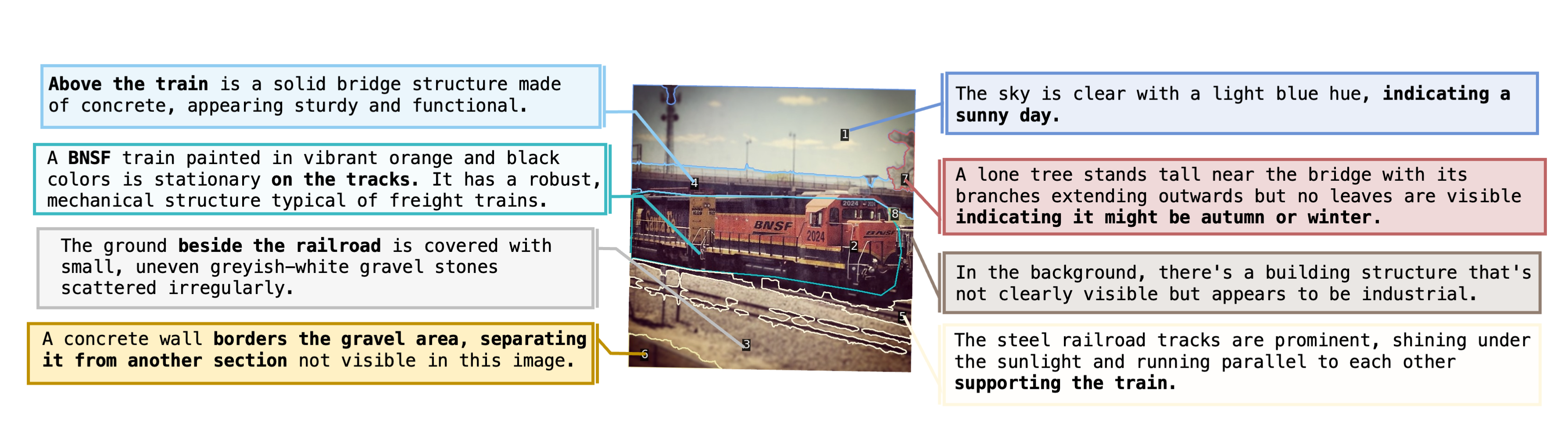} %
\caption{A visualization of Pix2Cap-COCO. Leveraging GPT-4V, our pixel-level captions contain comprehensive semantic information, including detailed object descriptions, interactions with surroundings, OCR information, and basic reasoning.}
\label{datasetvis}
\end{figure*}

To improve the diversity of the generated captions, we craft a series of prompts and randomly select one for each image. Details of the prompts can be found in the supplementary materials.

Finally, we re-annotate a subset of the collected data to create a high-quality validation set. During this process, annotators are presented with the marked images~\citep{yang2023set} and tasked with verifying the accuracy of the captions and their alignment with the visual information. For each annotation, annotators choose one of three options: (1) retain the captions without modification if they are correct, (2) correct any errors in the captions, or (3) skip the annotation if the caption is entirely inaccurate. Our annotation team, comprising 20 college students, re-annotated 15,281 instances across 2,338 images. To ensure quality, each caption underwent verification by at least two annotators. An overview of our dataset pipeline is provided in~\cref{fig:fdatasetpipeline}.

\subsection{Dataset Analysis}

We quantitatively compare Pix2Cap-COCO with existing region-level captioning datasets in~\cref{table:linguistic}.

\noindent \textbf{Data Scale:} Pix2Cap-COCO comprises 20,550 images sampled from the COCO dataset, with 167,254 detailed pixel-level captions. We partition it into a training set of 18,212 images and a validation set of 2,338 images. As can be seen from the overall statistics of~\cref{table:linguistic}(left), Pix2Cap-COCO provides significantly more object descriptions than existing datasets like RefCOCO~\citep{kazemzadeh2014referitgame} and RefCOCO+~\citep{yu2016modeling}. This is achieved through our design of an effective and automated data collection pipeline, which significantly enhances scalability compared to traditional, labor-intensive annotation methods. 

\noindent \textbf{Caption Length and Word Distribution}: in the right side of~\cref{table:linguistic} and Plot A of~\cref{fig:combine}, we also conduct a detailed statistical analysis of the captions in different datasets. The average caption length in Pix2Cap-COCO is 22.94 words, significantly surpassing the previous region-level datasets. Additionally, Pix2Cap-COCO demonstrates a balanced use of linguistic elements, averaging 7.08 nouns, 3.46 adjectives, and 3.42 verbs per caption, and achieves the highest average counts of descriptive elements per caption among comparable datasets. 

\cref{fig:combine} presents several more statistics of Pix2Cap-COCO by specifically assessing the distribution of descriptive words in the annotated captions, \eg, color, size, or shape. We can see that \(92\%\) of expressions in Pix2Cap-COCO contain more than one attribute, a significant increase compared to \(9\%\) in RefCOCO, which further demonstrates the richness and granularity of Pix2Cap-COCO. The distribution of common nouns, adjectives, and verbs in Pix2Cap-COCO captions is displayed in Plot C.

Beyond the intrinsic description of separate objects, their interactions are also crucial for fine-grained instance-level understanding. Therefore, our dataset construction pipeline tasks GPT-4V to explicitly depict the relationships between different objects to collect highly context-aware annotations. The captions in Pix2Cap-COCO effectively represent both basic positional relationships (\eg, next to, above) and more intricate spatial interactions (\eg, partially occluded by, vertically aligned with). We provide a visualization sample of Pix2Cap-COCO in~\cref{datasetvis}, you can refer to the supplementary material for more visualization results.

\definecolor{lightblue}{RGB}{173, 216, 230}
\definecolor{lightgreen}{RGB}{112, 205, 190}

\definecolor{cyellow}{RGB}{255, 221, 142}
\definecolor{cgreen}{RGB}{112, 205, 190}
\definecolor{cgray}{RGB}{132, 152, 171}
\definecolor{corange}{RGB}{245, 170, 97}

\begin{figure*}[t]
\centering
\includegraphics[width=0.9\linewidth]{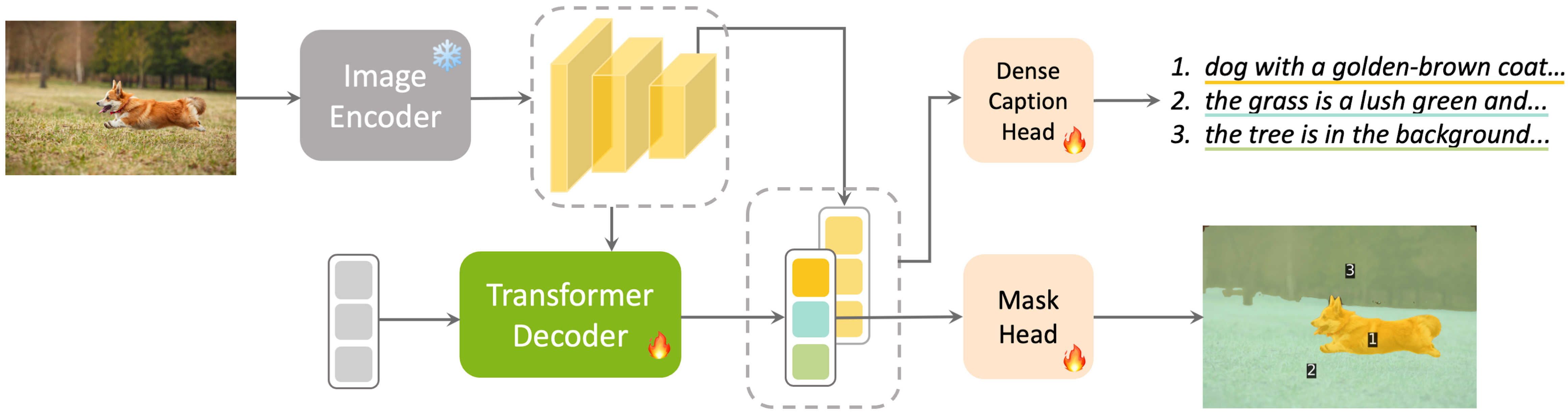}
\caption{An overview of our proposed model. Our proposed model follows an encoder-decoder architecture designed to perform the panoptic segmentation-captioning task end-to-end. Refer to the main text for details.}
\label{fig:framework}
\end{figure*}

\section{Panoptic Segmentation-Captioning}
\label{sec:Panoptic Segmentation-Captioning}

\subsection{Task Definition}
\label{sec:task}
Panoptic segmentation-captioning combines visual recognition and natural language generation challenges by requiring a model to both segment and describe every individual object in an image. Specifically, the expected outputs for this task are a list of mask-caption pairs: $\{(m_i,d_i)\}_{i=1}^{N}$, where \(m_i\) represents the mask of the instance \(i\) and \(d_i\) represents its corresponding pixel-level dense caption.

\noindent \textbf{Evaluation Metrics.} Following region-level dense captioning methods\citep{wu2022grit,johnson2016densecap}, we use mean Average Precision (mAP) to evaluate the overall performance of the model on the panoptic segmentation-captioning task. The mAP is calculated based on two sets of thresholds: Intersection over Union (IoU) thresholds for segmentation, set at 0.3, 0.4, 0.5, 0.6, and 0.7, and METEOR\citep{banerjee2005meteor} score thresholds for dense captioning, set at 0, 0.05, 0.1, 0.15, 0.2, and 0.25. The final mAP is determined by averaging the Average Precision (AP) across all combinations of this segmentation and captioning thresholds.

To further assess the quality of dense captions, we apply additional metrics commonly used in image captioning tasks~\citep{wang2022git,wang2022ofa,li2022mplug}, including BLEU~\citep{papineni2002bleu}, CIDEr~\citep{vedantam2015cider}, ROUGE~\citep{lin2004automatic}, and SPICE~\citep{anderson2016spice}, to the panoptic segmentation-captioning task. We define this metric as:

\[
m@\text{kIoU} = \frac{tp}{tp + fp + fn} \cdot m
\]

Where \( tp \), \( fp \), and \( fn \) represent the total counts of true positives, false positives, and false negatives, respectively, based on a segmentation IOU threshold \( k \), which is set at 0.5. The variable \( m \) represents the selected evaluation metrics from the list above.

\subsection{A Simple Baseline}
\label{sec:baseline}
We design a simple baseline for this task by appending a captioning head on X-Decoder~\citep{zou2023generalized}. X-Decoder is a versatile object segmentation framework that follows encoder-decoder architecture. It extracts multi-scale feature~\citep{zou2023generalized,cheng2021mask2former} through an image encoder to capture hierarchical visual information, and utilizes a transformer decoder to enable each object token to learn the object-specific feature. Subsequently, a multi-layer perceptron (MLP) is employed as the mask head to generate pixel-level segmentation masks by decoding the object tokens.

To generate captions for the segmented instances, we attach a dense caption head to the X-Decoder, which consists of 6 transformer layers. We pool the last layer of image features as global tokens and concatenate them with object tokens to enrich the decoder with broader contextual information. Within the dense caption head, causal attention is applied to ensure the caption is generated sequentially. Each text token attends to both preceding text tokens and concatenated tokens from the transformer decoder, capturing specific object features and their contextual interactions. We name our baseline model Pix2Cap to facilitate reference. Cross-entropy with a label smoothing of 0.1 is used as the loss function~\citep{wu2022grit} for pixel-level dense caption generation:

\[\mathcal{L}_{\text{caption}} = \frac{1}{N+1} \sum_{i=1}^{N+1} \text{CE}(y_i, \tilde{y}_i)\]

where \( \tilde{y}_i = p(y_i | \mathbf{o}, \{y_j\}_{j=0}^{i-1}) \) represents the predicted score for the \(i\)-th text token \(y_i\), given the output \( \mathbf{o} \) from transformer decoder and all previously generated text tokens \( \{y_j\}_{j=0}^{i-1} \). \(N\) denotes the total number of text tokens in the sequence. The class loss (\(\mathcal{L}_{\text{class}}\)) is defined as the binary cross-entropy between the class label and the dot-product of the object tokens and the concept embeddings, following the X-Decoder.
We use a combination of binary cross-entropy and dice loss as the mask loss (\(\mathcal{L}_{\text{mask}}\)). Finally, the model can be trained in an end-to-end manner using the following training objective:

\[
\mathcal{L} = \lambda_{\text{caption}}\mathcal{L}_{\text{caption}} + \lambda_{\text{mask}}\mathcal{L}_{\text{mask}} + \lambda_{\text{class}}\mathcal{L}_{\text{class}}
\]

\(\lambda_{\text{caption}}\), \(\lambda_{\text{mask}}\), and \(\lambda_{\text{class}}\) are set to 0.1, 5.0, and 2.0, respectively. Hungarian matching~\citep{carion2020end, cheng2021per} is used to find the allocation with the lowest cost. During inference, we employ beam search~\citep{brown1993mathematics} following \citep{wu2022grit}. 

\begin{table*}[t!]
    \centering
    \caption{Qualitative results of our proposed Pix2Cap model. Our model can simultaneously produce detailed, dense captions and high-quality panoptic segmentation. The caption related to each mask is not only an appearance description of the object but also includes its interaction with its surroundings. Captions of the same category are marked with the same color. Zoom in for more details.}
    \label{tab:panoptic_caption_qulitative}
    \resizebox{\textwidth}{!}{
    \footnotesize
    \begin{tabular}{p{6cm} p{6cm} p{21.3cm}}
        \toprule
        \multicolumn{1}{c}{\fontsize{17}{19}\selectfont \textbf{Image}} & \multicolumn{1}{c}{\fontsize{17}{19}\selectfont \textbf{Segmentation}} & \multicolumn{1}{c}{\fontsize{17}{19}\selectfont \textbf{Dense Captions}} \\

        \midrule

        \makecell{\includegraphics[width=6cm]{vis/000000177893.jpg}} &
        \makecell{\includegraphics[width=6cm]{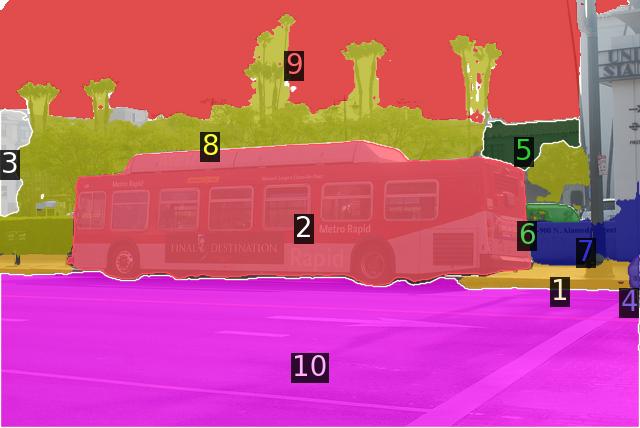}} &
        \makecell[l]{ %
\colorbox{cyellow}{\Large1:the pavement is made of concrete slabs arranged in an orderly pattern beside the road where the bus is} \\
\colorbox{cyellow}{\Large stationed.}\\
\colorbox{blue!30}{\Large2:a red and white bus with the displayed on the front, parked on the road. it has a rectangular shape and}\\
\colorbox{blue!30}{\Large is located in the center of the image.}\\
\colorbox{orange!50}{\Large3:buildings are partially visible behind the trees ; they have a traditional architectural style with brick}\\
\colorbox{orange!50}{\Large facades.}\\
\colorbox{blue!30}{\Large4:a car is partially visible behind the bus ; it's white and seems to be of a compact size based on its} \\
\colorbox{blue!30}{\Large proportion relative to other objects.}\\
\colorbox{orange!50}{\Large5:a house-like structure can be seen behind the bus ; it's white and larger compared to other buildings but}\\
\colorbox{orange!50}{\Large details are not clear due to distance.}\\
\colorbox{blue!30}{\Large6:a car is partially visible behind the bus ; it's white and seems to be of a compact size based on its} \\
\colorbox{blue!30}{\Large proportion relative to other objects.}\\
\colorbox{gray!30}{\Large7:a wall or barrier is partially visible behind the bus, it's color and texture are not clearly distinguishable}\\
\colorbox{gray!30}{\Large due to obstructions.}\\
\colorbox{lightgreen}{\Large8:a tree with green leaves is partially visible to the left of the image, its branches extending upwards } \\
\colorbox{lightgreen}{\Large and outwards.}\\
\colorbox{white}{\Large9:the sky is partly cloudy with a mix of white clouds and blue sky visible.}\\
\colorbox{cyellow}{\Large10:the road is a flat, smooth surface with a greyish color, extending from the foreground to the }\\
\colorbox{cyellow}{\Large background of the image.}\\
        } \\

\midrule
        \makecell{\includegraphics[width=6cm]{vis/000000090003.jpg}} &
        \makecell{\includegraphics[width=6cm]{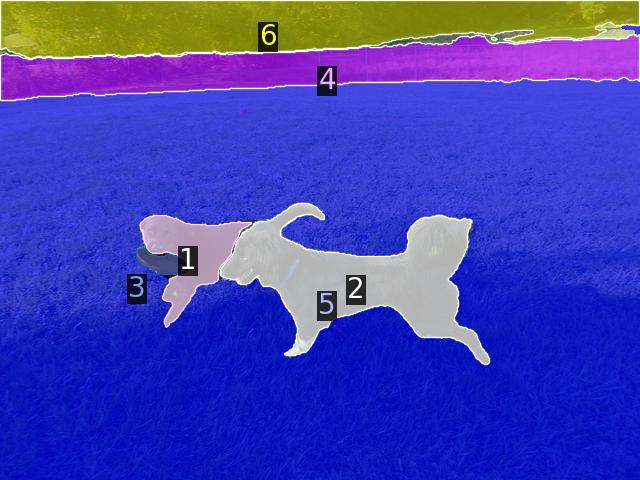}} &
        \makecell[l]{\colorbox{cgray!60}{\Large1:a black dog with a shiny coat is in motion, its body language suggesting it's engaged in play or}\\
        \colorbox{cgray!60}{\Large reacting to something.}\\
\colorbox{cgray!70}{\Large2:a black dog with a shiny coat is in motion, its body language suggesting it's engaged in play.}\\
\colorbox{white}{\Large3:a red frisbee is captured mid-air close to the dog's mouth, indicating that it might be caught by the dog.}\\
\colorbox{gray!30}{\Large4:a chain-link fence is visible in the background, separating the grassy foreground from the tree-filled} \\
\colorbox{gray!30}{\Large background.}\\
\colorbox{white}{\Large5:the grass is a lush green and appears well-maintained, covering the ground uniformly.}\\
\colorbox{lightgreen}{\Large6:trees in the background are tall with dense foliage, casting shadows below them.}\\       
} \\

\midrule

        \makecell{\includegraphics[width=6cm]{vis/000000031217.jpg}} &
        \makecell{\includegraphics[width=6cm]{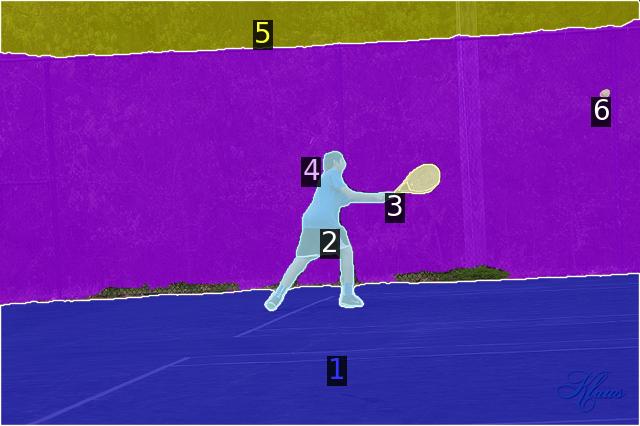}} &
        \makecell[l]{
        \colorbox{cyellow}{\Large1:field is a flat, smooth surface with a greenish hue, marked with white lines that define the boundaries}\\     
        \colorbox{cyellow}{\Large and center of the tennis court. it is surrounded by a dark chain-link fence that}\\
        \colorbox{white}{\Large2:a person is captured mid-motion, wearing a blue top and dark shorts, holding a tennis racket. }\\
        \colorbox{white}{\Large3:the person is holding a tennis racket in their right hand, ready to swing or having just swung. }\\
        \colorbox{gray!30}{\Large4:the fence is made of a dark, tightly woven mesh material, providing a barrier around the tennis court. }\\
        \colorbox{gray!30}{\Large it appears sturdy and is partially visible behind the person and the playing field. }\\
        \colorbox{lightgreen}{\Large5:trees are visible in the background behind the fence, they are lush and provide a natural backdrop to}\\
        \colorbox{lightgreen}{\Large the tennis court.}\\ 
        \colorbox{white}{\Large6:a small yellow tennis ball is captured in motion near the player's racket.}} \\
        \bottomrule
    \end{tabular}
} 
\end{table*}

\begin{table}[!ht] 
\centering
\caption{Quantitative results of our model on the panoptic segmentation-captioning task on Pix2Cap-COCO benchmark.}
\label{tab:panoptic_caption_quantify}
\vspace{-0.05in}
\setlength{\tabcolsep}{1.mm} 
\footnotesize
\begin{tabular}{c|*{5}{c}}
\toprule
& \textbf{mAP}  & \textbf{BLEU@4}  &  \textbf{CIDEr}  &  \textbf{ROUGE}  &  \textbf{SPICE}  \\
\midrule
GRiT+Centermask &16.3 &5.0 &30.6 &14.2 &10.7 \\
Pix2Cap &\textbf{17.8} &\textbf{5.4} &\textbf{32.7} &\textbf{15.8} &\textbf{11.3}    \\
\bottomrule
\end{tabular}
\end{table}

\subsection{Experimental Results}
\label{sec:implement}
\textbf{Implementation Details:} we train Pix2Cap (Sec.~\ref{sec:baseline}) on the training set of Pix2Cap-COCO with the input size of \(1024\times1024\) for 25 epochs with AdamW optimizer~\citep{loshchilov2017decoupled}. The batch size is 32, and the initial learning rate is set to \(10^{-5}\) with a weight decay of 0.05.  We initialize our model with the weight of X-Decoder~\citep{zou2023generalized}. The image encoder is frozen while the transformer decoder, dense caption head, and mask head are trainable. The experiments are conducted on 8 NVIDIA A100 GPUs with 80G memory.

\begin{table}[h] 
\centering
\caption{Comparison of our model with recent panoptic segmentation methods on COCO val2017. }
\label{tab:xdecoder}
\vspace{-0.1in}
\setlength{\tabcolsep}{3.5mm}
\renewcommand{\arraystretch}{1.1} 
\footnotesize
\begin{tabular}{c|*{3}{c}}
\toprule
& \textbf{PQ}  & \textbf{PQ$_{th}$}  &  \textbf{PQ$_{st}$}  \\
\midrule
MaskFormer \citep{cheng2021per}     &52.7 &58.5 &44.0\\
K-Net \citep{zhang2021knet}    &54.6 &60.2 &46.0  \\
Panoptic SegFormer \citep{li2021panoptic}   &55.8 &61.7 &46.9 \\
X-Decoder \citep{zou2023generalized}     &56.7 &62.9 &47.3 \\
\midrule
\textbf{X-Decoder(Ours)}       
&\textbf{57.0} 
&\textbf{63.2} 
&\textbf{47.5}    \\
\bottomrule
\end{tabular}
\end{table}

\label{sec:results}
\begin{table*}[ht!]
\centering
\caption{Comparision with existing instance-level large multimodal models on ViP-Bench.}
\label{tab:vip}
\vspace{-0.05in}
\setlength{\tabcolsep}{2mm} 
\footnotesize
\begin{tabular}{lccccccc}
\toprule
\textbf{Model}  & \textbf{Rec} & \textbf{OCR} & \textbf{Know} & \textbf{Math} & \textbf{Rel} & \textbf{Lang} & \textbf{All} \\
\midrule
InstructBLIP-7B \citep{instructblip}  & 36.9 & 16.3 & 34.2 & 22.3 & 26.8 & 7.5 & 31.7 \\
InstructBLIP-13B \citep{instructblip} & 42.5 & 12.2 & 37.5 & 3.2 & 33.2 & 12.5 & 35.8 \\
Shikra-7B \citep{chen2023shikra} & 40.2 & 10.0 & 28.0 & 3.5 & 18.9 & 20.6 & 33.7 \\
Kosmos-2 \citep{kosmos-2} & 29.5 & 14.2 & 18.5 & 9.7 & 7.5 & 21.9 & 26.9 \\
Qwen-VL-Chat \citep{Qwen-VL}  & 43.0 & 30.4 & 40.2 & 9.7 & 25.7 & 28.7 & 39.2 \\
GPT4RoI-7B \citep{zhang2023gpt4roi} & 35.6 & 16.7 & 29.7 & 9.7 & 32.5 & 13.8 & 35.1 \\
\midrule
\textbf{GPT4RoI-7B(Ours)} & 
\textbf{46.8\textcolor{green!50!black}{(+11.2 $\uparrow$)}} & 
\textbf{17.5\textcolor{green!50!black}{(+0.8 $\uparrow$)}} & 
\textbf{31.4\textcolor{green!50!black}{(+1.7 $\uparrow$)}} & 
\textbf{12.9\textcolor{green!50!black}{(+3.2 $\uparrow$)}} & 
\textbf{33.9\textcolor{green!50!black}{(+1.4 $\uparrow$)}} & 
\textbf{35.6\textcolor{green!50!black}{(+22.2 $\uparrow$)}} & 
\textbf{40.2\textcolor{green!50!black}{(+5.1 $\uparrow$)}} \\

        \bottomrule
    \end{tabular}
\end{table*}

\noindent \textbf{Main Results}: The panoptic segmentation-captioning performance is evaluated on the validation set of Pix2Cap-COCO. We design another baseline for better comparison by using GRiT~\citep{wu2022grit} to detect bounding boxes and captions first, and CenterMask~\citep{lee2020centermask} to segment the masks on the bounding boxes (denoted as GRiT+CenterMask, please refer to the supplementary materials for details). The results are compared in~\cref{tab:panoptic_caption_quantify}, from which we can see that Pix2Cap can achieve competitive captioning performance on our Pix2Cap-COCO, \eg, 32.7 CIDER, consistently better than the GRiT+CenterMask baseline on all metrics.

We also present qualitative results in~\cref{tab:panoptic_caption_qulitative} for a better illustration. Pix2Cap owns the capability to generate both high-quality panoptic segmentation masks and detailed captions, based on their visual appearance and context. For example, the function description for the fence in row 2 is ``separating the grassy foreground from the tree-filled background'' while in row 3 it is ``providing a barrier around the tennis court'', demonstrating a strong interaction between the instance and its surroundings.

\noindent \textbf{Can caption help panoptic segmentation?} Since Pix2Cap-COCO can provide detailed captions for the instances, it is interesting to see if they can help to improve the panoptic segmentation performance. We evaluate our Pix2Cap on the panoptic COCO val2017. As shown in~\cref{tab:xdecoder}, the integration of dense caption training alongside standard panoptic mask training can lead to improvements in PQ, \( \text{PQ}_{\text{th}} \), and \( \text{PQ}_{\text{st}} \) scores compared to the original X-Decoder. The results demonstrate the potential of leveraging dense captions as complementary supervision to enhance panoptic segmentation performance. Details of this part can be found in the supplementary material.

\subsection{Pix2Cap-COCO for Large Multimodal Models}

\begin{table}[t!] 
\centering
\footnotesize
\caption{Quantitative comparison on the test set of Visual Genome. }
\vspace{-0.05in}
\label{tab:VG}
\setlength{\tabcolsep}{1.2mm} 
\renewcommand{\arraystretch}{1.1}
\begin{tabular}{c|*{4}{c}}
\toprule
& \textbf{CIDEr}   & \textbf{ROUGE}  &\textbf{SPICE}  \\
\midrule
GRiT \citep{wu2022grit}  & 142.0 & - & - \\
Shikra-7B \citep{chen2023shikra} & 115.8 & 30.3 & -\\
GPT4RoI-7B \citep{zhang2023gpt4roi}  &154.9 &36.8 & 33.8 \\
\midrule
\textbf{GPT4RoI-7B (Ours)} & \textbf{156.3\textcolor{green!50!black}{(+1.4 $\uparrow$)}}
& \textbf{37.2\textcolor{green!50!black}{(+0.4 $\uparrow$)}} 
& \textbf{34.3\textcolor{green!50!black}{(+0.5 $\uparrow$)}}\\
\bottomrule
\end{tabular}
\end{table}

In this section, we conduct experiments to demonstrate the benefits of our Pix2Cap-COCO in improving the instance-level understanding capability of existing large multimodal models (LMMs). We choose GPT4RoI\citep{zhang2023gpt4roi} as the baseline in this section, which incorporates spatial instruction tuning with region-level caption datasets, significantly boosting LMM performance in diverse region understanding tasks like region captioning.

\begin{figure*}[th]
\centering
\includegraphics[width=0.95\linewidth]{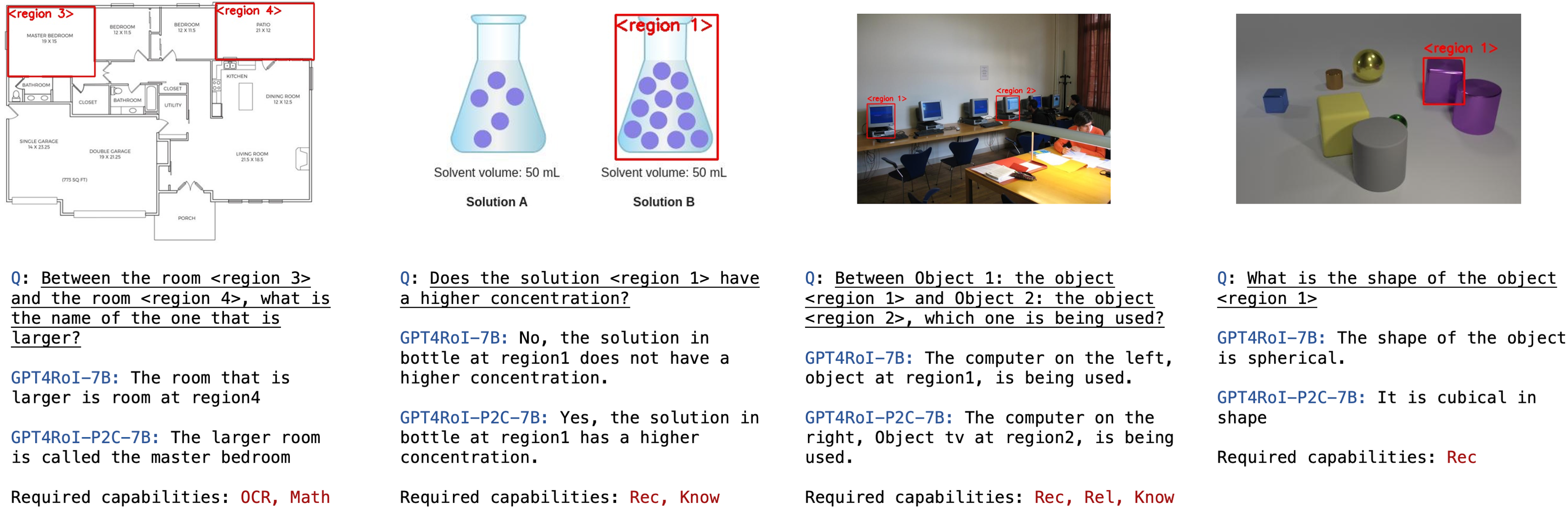}
\caption{Visual comparison on ViP-Bench between GPT4RoI-7B and GPT4RoI-P2C-7B (\textit{P2C} denotes training with Pix2Cap-COCO). }
\label{fig:vip_vis}
\end{figure*}

The training process for GPT4RoI\citep{zhang2023gpt4roi} consists of two stages. In the first stage, the model is trained on datasets such as COCO\citep{lin2014microsoft}, RefCOCO\citep{kazemzadeh2014referitgame}, and RefCOCO+\citep{yu2016modeling}, equipping it with foundational knowledge of object categories and attributes. In the second stage, it is further finetuned on more complex datasets, including RefCOCOg\citep{yu2016modeling}, Visual Genome (VG)~\citep{krishna2017visual},  Flicker30k~\citep{plummer2015flickr30k}, LLaVA150k~\citep{liu2024visual}, and VCR~\citep{zellers2019recognition} to enhance its ability to tackle complex region understanding tasks. We retrain GPT4RoI-7B in stage 2 using Pix2Cap-COCO alongside other datasets follow the original recipe of GPT4RoI and evaluate on ViP-Bench~\citep{cai2024vipllava} and Visual Genome~\citep{krishna2017visual}.

ViP-Bench\citep{cai2024vipllava} provides a comprehensive assessment for multi-modal models across six aspects: recognition, OCR, knowledge, math, relationships, and language generation. As shown in~\cref{tab:vip} and~\cref{fig:vip_vis}, incorporating Pix2Cap-COCO during the supervised fine-tuning (SFT) stage significantly boosts the performance of GPT4RoI, with an overall improvement of \(5.1\%\). Notably, the recognition and language generation scores of GPT4RoI are increased by \(11.2\%\) and \(22.2\%\), respectively, highlighting the effectiveness of Pix2Cap-COCO in providing rich and descriptive annotations that bridge visual recognition and language generation tasks. We provide more visualization results in the supplementary material. 

We also evaluate our model on VG~\citep{krishna2017visual}, and present the quantitative comparison with existing instance LMMs in~\cref{tab:VG}. Compared to the original GPT4RoI, our model achieves \(1.4\%\), \(0.4\%\), and \(0.5\%\) gains on the CIDEr, ROUGE, and SPICE metrics, respectively. This indicates that the fine-grained captions in our dataset could enable models to better capture the inter-object. 

\vspace{-0.05in}
\section{Conclusion}
\label{sec:conc}
\vspace{-0.05in}
In this paper, we introduced Pix2Cap-COCO, the first panoptic pixel-level caption dataset designed to advance fine-grained visual comprehension. Pix2Cap-COCO overcomes the limitations of existing segmentation or caption datasets by providing detailed, object-specific captions precisely aligned with segmentation masks. To construct Pix2Cap-COCO in a cost-effective and scalable way, we developed an automated pipeline that marks instances within images with Set-of-Mark and employs GPT-4V to generate corresponding descriptions. Building on Pix2Cap-COCO, we proposed a novel and challenging task, \ie, panoptic segmentation-captioning, which integrates the prediction of segmentation masks and detailed captions. Extensive experiments across multiple benchmarks demonstrated that Pix2Cap-COCO could not only establish new standards for instance-level understanding but also significantly enhance the performance of current multimodal models in fine-grained visual tasks.

\bibliographystyle{ieeenat_fullname}
\bibliography{main}

\clearpage
\setcounter{page}{1}
\section{Additional Dataset Visualizations}
\label{sec:dataset_vis}
We provide additional visualizations of our dataset in this section. As illustrated in \cref{tab:dataset_qulitative}, our captions are highly detailed and explicitly capture interactions between objects. Moreover, the pixel-level object descriptions enable our dataset to effectively differentiate between visually similar objects (\eg, the oranges in row 1 and the cows in row 2). Unlike image-level and region-level caption datasets, our approach ensures precise alignment between visual inputs and captions, avoiding the common issues of misalignment seen in other datasets.

\section{Details on Prompt Engineering}
\label{sec:prompt}
Prompt engineering plays a critical role in guiding GPT-4V to generate detailed and structured pixel-level captions for our dataset. We present two prompts we mainly used in \cref{fig:instruction_format}. To ensure clarity and accuracy, we define the task first, specify the labeled objects and their categories, and emphasize the need for unique and detailed descriptions for each object. Instructions include capturing comprehensive details such as color, shape, texture, motion, and relative positions, along with interactions between objects. Explicit formatting guidelines are provided to maintain uniformity, and constraints are placed on retaining given categories and avoiding grouped descriptions.
\section{Additional Implementation Details}
\label{sec:implementation}
\subsection{GRiT for Panoptic Segmentation-Captioning}

GRiT\citep{wu2022grit} is a robust model designed for open-vocabulary object detection and region-level dense captioning. It inherits the open-set feature of generative methods by introducing an additional language model as the captioning head. To adapt GRiT for panoptic segmentation-captioning, we enhance it with CenterMask\citep{lee2020centermask}, integrated as a mask head atop the decoder outputs. The mask head processes RoIAligned features for each proposed object and predicts a binary mask for each. Following \citep{carion2020end}, an argmax operation is applied to the mask scores at each pixel, assigning categories to the masks and ensuring no overlaps. The mask head is trained jointly with all other components on the Pix2Cap-COCO dataset. To optimize performance, we perform a grid search to tune hyper-parameters like the learning rate and the weight of the mask loss. The model is trained on 8 NVIDIA A100 GPUs with the mask loss weight set to 1 and a batch size of 32 for 180,000 iterations. The learning rate is initialized at \(8 \times 10^{-5}\).

\subsection{Details on Panoptic Segmentation}
We examine whether detailed pixel-level captions could enhance segmentation performance in our main paper. During the dataset construction phase, we removed annotations with low-quality captions, resulting in our dataset not fully overlapping with the original COCO. To address this, we supplement the missing data by reintroducing annotations that had been filtered out during our dataset selection process, assigning the category name of each annotation as its caption while preserving other attributes. We then train our baseline model mentioned in \cref{sec:baseline} on this padded dataset, using the same recipe in \cref{sec:implement} for 50 epochs. Since the architecture of our model effectively functions as an X-Decoder with an added dense captioning head, by retraining this model on our dataset, we can directly compare it with the original X-Decoder to assess the impact of pixel-level dense captioning on segmentation performance.

\subsection{Details on ViP-Bench}
ViP-Bench\citep{cai2024vipllava} is a comprehensive benchmark designed to assess the reasoning and interpretative abilities of multimodal models using 303 unique image-question pairs. Each pair combines an image with a diverse visual reasoning question, challenging the model's capacity for region-level visual understanding. The benchmark focuses on six critical aspects: recognition, Optical Character Recognition (OCR), knowledge integration, mathematical reasoning, object relationship comprehension, and language generation. Responses from multimodal models are compared against human-annotated answers and evaluated by GPT-4, which assigns scores on a scale of 0 to 10. This scoring system provides a standardized and quantitative measure for comparing model performance, offering a robust benchmark for assessing their ability to process and interpret complex visual data. We provide additional qualitative results on ViP-Bench in  ~\cref{tab:vis_vip}. 

\section{Qualitative Results and Failure Case }
In this section, we provide additional qualitative results of our Pix2Cap model. As illustrated in \cref{tab:additional_dataset_qulitative}, Pix2Cap establishes a solid baseline for the panoptic segmentation-captioning . However, in complex scenarios, it occasionally generates inaccurate object attributes (\eg, the location of the brick wall in row 5). This issue likely arises from the limited size of the object tokens (\(\mathbf{N} \times 512\)), which may lead to insufficient information allocation when processing scenes with numerous objects. We plan to explore this issue further in our future work.

\begin{table*}[t!]
    \centering
    \caption{Additional visualizations of our proposed Pix2Cap-COCO dataset.}
    \label{tab:dataset_qulitative}
    \resizebox{\textwidth}{!}{
    \begin{tabular}{p{6cm} p{14cm}} 
        \toprule
        \multicolumn{1}{c}{\fontsize{13}{15}\selectfont \textbf{Image}} & 

        \multicolumn{1}{c}{\fontsize{13}{15}\selectfont \textbf{Dense Captions}} \\

        \midrule

        \makecell{\includegraphics[width=6cm]{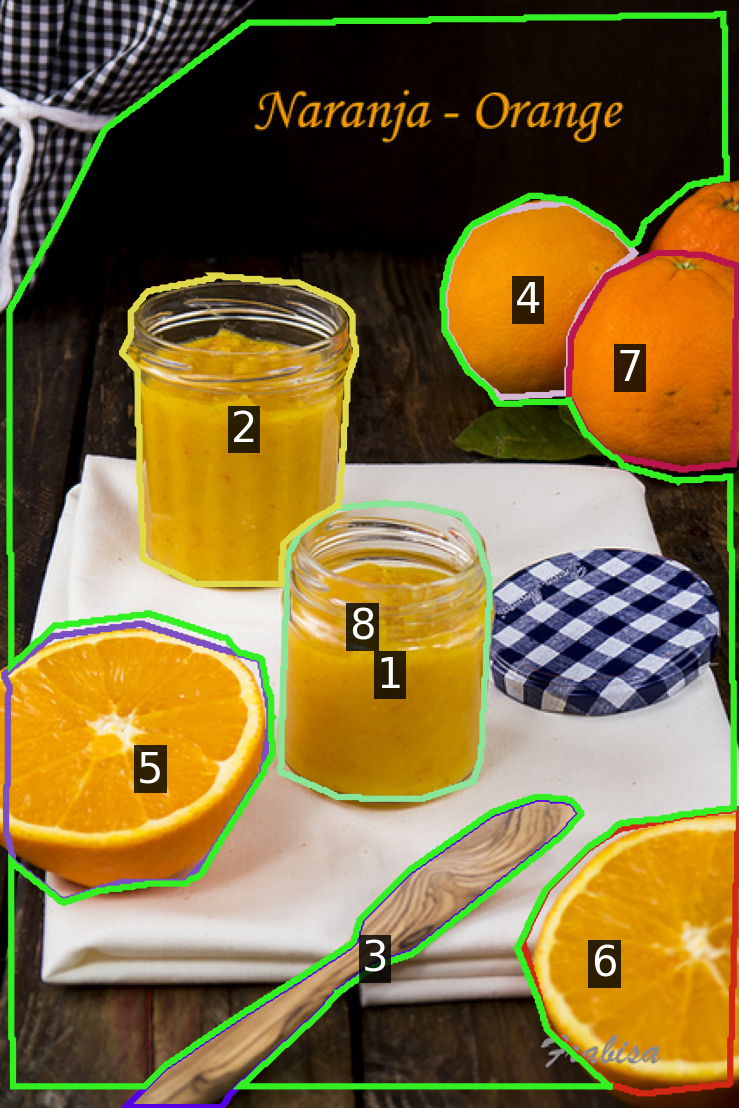}} &

        \makecell[l]{
1:Another glass cup filled with orange jam or marmalade but slightly smaller in size.\\
2:A glass cup filled with orange jam or marmalade, it has an open top and is placed to the left\\
side on the table.\\
3:A wooden-handled knife rests on the table close to a sliced piece of orange.\\
4:Positioned next to this whole uncut orange has a bright color indicating ripeness.\\
5:This is a half-sliced orange with juicy pulp visible, placed on the white cloth of the dining table.\\
6:A juicy slice of an orange that lies flat on the table near the knife.\\
7:A whole uncut orange sitting next to another one, both are positioned at the top right corner of\\
the image.\\
8:The dining table is covered with a white cloth, and various items are placed on it, including\\
cups of orange jam, slices of oranges, and a knife.\\
        } \\
                \midrule
        \makecell{\includegraphics[width=6cm]{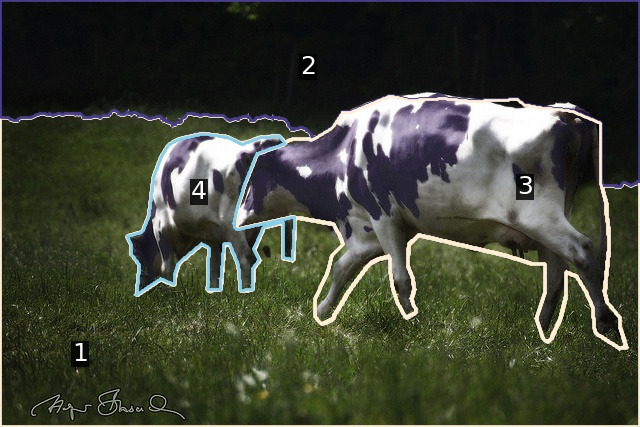}} &

        \makecell[l]{
1:The grass is lush and green , covering the ground uniformly. It appears well-maintained \\
and provides a natural base for the other objects in the image.\\
2:The trees are in the background, their outlines slightly blurred but still visible. They stand tall\\
and provide a contrasting dark green backdrop to the bright foreground.\\
3:This cow is larger, with a white body adorned with large black spots. It's standing upright and\\
appears healthy and well-fed.\\
4:This smaller cow has similar coloring to it but is distinguished by its size and posture - it's \\
head is down, suggesting it might be grazing.\\
        } \\
        \midrule

        \makecell{\includegraphics[width=6cm]{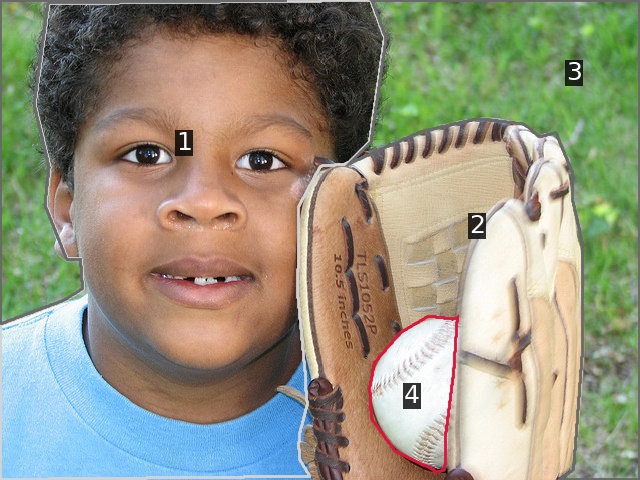}} &

        \makecell[l]{
1:A child is visible from the chest up, wearing a light blue shirt. The child has curly hair and a\\
cheerful expression, with eyes looking towards something interesting.\\
2:The glove is tan and well-worn, with dark brown lacing. It's open and appears to be in the act\\
of catching a ball.\\
3:The background consists of vibrant green grass illuminated by natural light, providing a fresh\\
and open atmosphere.\\
4:A white baseball with brown stitching is partially inside the baseball glove, appearing as if it\\
has just been caught.\\
        } \\
        \midrule
        \makecell{\includegraphics[width=6cm]{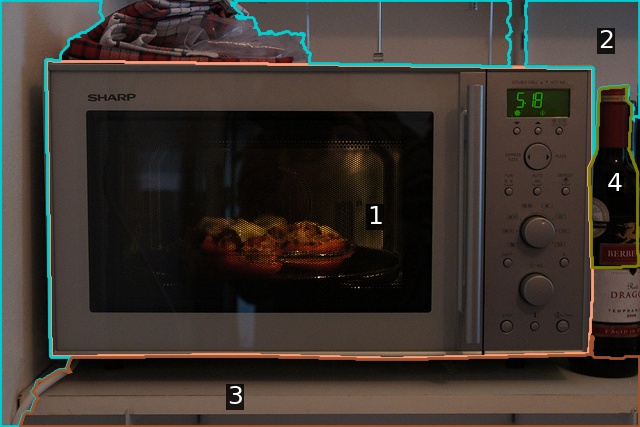}} &

        \makecell[l]{
1:The microwave is a SHARP brand, with a grey exterior and a dark interior visible through\\
the glass door. It has a digital display.\\
2:The wall behind the microwave is painted in a light color, possibly white or cream, with some\\
 items like clothes hanging on it.\\
3:A flat surface, possibly a countertop or table, in light color supports the microwave. It extends\\
across the bottom of the image.\\
4:A dark bottle with a label is visible to the right of the microwave, standing on the same surface\\
as the microwave.\\
        } \\
        \bottomrule
    \end{tabular}
    } 
\end{table*}

\begin{figure*}[t]
    \centering
    \begin{tcolorbox}[colframe=black, colback=gray!10, rounded corners, width=\textwidth]
        \textbf{PROMPT1} = ``You are an AI visual assistant that can analyze a single image. In the given image, I label \texttt{\(<\)num instances\(>\)}  objects (or background) by marking each with a bright numeric ID at the center and its boundary. I could also tell you their categories: \texttt{\(<\)img categories\(>\)}.

Note that I want to use your description to create a grounding dataset, therefore, your descriptions for different objects should be unique, \ie, if multiple objects of the same categories appear in the same image, your descriptions should be detailed enough to differentiate them.

Please describe the appearance of different visual objects and the interaction between them. Your description should be as detailed as possible, including details like color, shape, texture, emotion, motion, intention, object counts, the position of the objects, relative position between the objects. 

Your answer should contain details, and follow the following format:

\('''\)

The objects include: 

object\(\_\)id. summary of the object: description of the object (\eg, 1. Person: the person is wearing a pink jacket, black pants, and a pink beanie. They are holding ski poles and are skiing on a snowy mountain.);

object\(\_\)id. summary of the object: description of the object;

The interactions include:

object\(\_\)id and object\(\_\)id: interaction between them (\eg, 1 and 2: the person is skiing on the mountain using the skis.);

object\(\_\)id and object\(\_\)id: interaction between them;

\('''\)

Please pay attention to the categories of these objects and don't change them. Also, keep in mind that you should not group the objects, \eg, 2-5. people: xxx, be sure to describe each object separately (one by one). Please start your answer:"

\textbf{PROMPT2} = ``I have labeled \texttt{\(<\)num instances\(>\)} objects (or background) in this image and each with a bright numeric ID at the center and its boundary. Their categories are: \texttt{\(<\)img categories\(>\)}.

Please describe the appearance of different visual objects and the interaction between them. Your description should be as detailed as possible, including details like color, shape, texture, emotion, motion, intention, object counts, the position of the objects, relative position between the objects. 

Note that I want to use your description to create a grounding dataset, therefore, your descriptions for different objects should be unique, \ie, if multiple objects of the same categories appear in the same image, your descriptions should be detailed enough to differentiate them.

Your answer should contain details, and follow the following format:

\('''\)

The objects include: 

object\(\_\)id. summary of the object: description of the object (\eg, 1. Person: the person is wearing a pink jacket, black pants, and a pink beanie. They are holding ski poles and are skiing on a snowy mountain.);

object\(\_\)id. summary of the object: description of the object;

The interactions include:

object\(\_\)id and object\(\_\)id: interaction between them (\eg, 1 and 2: the person is skiing on the mountain using the skis.);

object\(\_\)id and object\(\_\)id: interaction between them;

\('''\)

Please pay attention to the categories of these objects and don't change them. Also, keep in mind that you should not group the objects, \eg, 2-5. people: xxx, be sure to describe each object separately (one by one). Please start your answer:"

    \end{tcolorbox}
    \caption{Instruction format for guiding GPT-4V to generate pixel-level captions.}
    \label{fig:instruction_format}
\end{figure*}
\begin{table*}[t!]

    \centering
    \caption{ Additional qualitative results of our proposed Pix2Cap model on panoptic segmentation-captioning task.}
    \label{tab:additional_dataset_qulitative}
    \large
    \resizebox{\textwidth}{!}{
    \begin{tabular}{p{6cm} p{6cm} p{16.5cm}} 
        \toprule
        \multicolumn{1}{c}{\fontsize{13}{15}\selectfont \textbf{Image}} & 
        \multicolumn{1}{c}{\fontsize{13}{15}\selectfont \textbf{Segmentation}} & 
        \multicolumn{1}{c}{\fontsize{13}{15}\selectfont \textbf{Dense Captions}} \\
        \midrule
        \makecell{\includegraphics[width=6cm]{vis/000000572900.jpg}} &
        \makecell{\includegraphics[width=6cm]{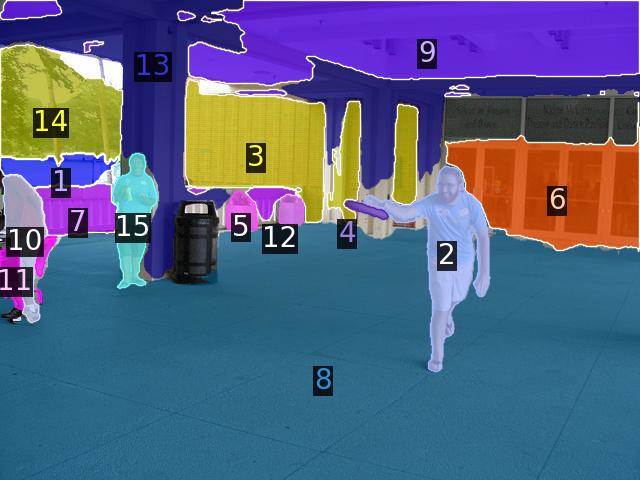}} &
        \makecell[l]{1:green grass patches are visible near the pavement, adding natural elements to this urban setting.\\
        2:a person in a blue t-shirt and grey shorts is captured mid-motion, appearing to be throwing\\
        something with their right hand.\\
        3:a section of brick wall is visible to the left, with a reddish-brown color and rough texture.\\
        4:a frisbee captured mid-air close to being caught by person.\\
5:a potted plant sits atop a ledge near the stairs. its leaves are vibrant green, and it adds a touch \\
of nature to the urban setting.\\
6:a closed door with a dark frame and light panels is visible on the left side of the image.\\
7:a white picket fence runs along beside where people are standing separating them from grassy\\
area.\\
8:the floor is made of square, light-colored tiles that are neatly arranged and cover the entire \\
room's base.\\
9:the ceiling is white and appears smooth, extending across the top of the image.\\
10:a person wearing a red top and black pants is standing on the pavement, facing away from\\
 the camera.\\
11:a chair made of metal and plastic material placed on pavement near stairs.\\
12:a potted plant sits atop a ledge near the stairs. its leaves are vibrant green, and it adds a touch \\
of nature to the urban setting.\\
13:the wall is a large, flat surface with a beige color and smooth texture. it extends across the \\
background of the image, providing a backdrop for other objects.\\
14:a tree with green leaves is visible behind the fence, partially obscured but adding natural \\
elements to the urban setting.\\
15:another person in blue attire appears engaged in an activity near the bench.\\
} \\

\midrule

        \makecell{\includegraphics[width=6cm]{vis/000000325991.jpg}} &
        \makecell{\includegraphics[width=6cm]{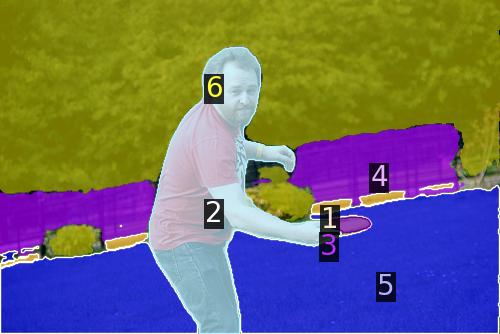}} &
        \makecell[l]{1:a grey pavement is visible at the bottom of the image, appearing solid and smooth.\\
2:a person is captured mid-motion, wearing a red tank top and dark shorts, appearing to be \\
throwing a frisbee.\\
3:a white frisbee is captured mid-air close to the person's outstretched hand.\\
4:a white fence is visible in the background, appearing sturdy yet unobtrusive, marking the\\
boundary of the yard.\\
5:the grass is a vibrant green and appears well-maintained, covering the ground uniformly.\\
6:the trees are lush and green, providing a natural backdrop to the scene. they are dense and\\
appear healthy, indicating a well-maintained environment.\\
} \\

\midrule

        \makecell{\includegraphics[width=6cm]{vis/000000573258.jpg}} &
        \makecell{\includegraphics[width=6cm]{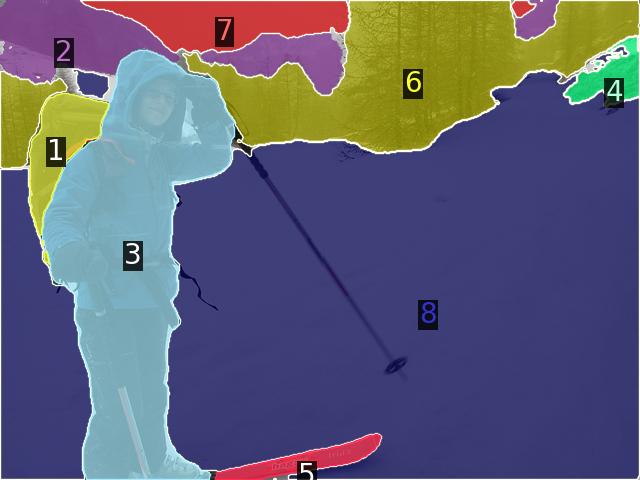}} &
        \makecell[l]{1:the person wears a black backpack with white straps. the backpack rests comfortably on their\\
        back, suggesting it might be prepared for outdoor activities.\\
2:the mountains are covered in snow, with their peaks appearing rugged and majestic against\\
the clear sky.\\
3:a person is wearing a blue jacket, black pants, sunglasses, and a beanie. they are holding onto\\
ski poles.\\
4:rocks are visible in the background, partially covered by snow and appearing rugged and steep.\\
5:the skis are long and narrow, appearing light in color as they glide across the surface of \\
the snow.\\
6:a group of bare trees with thin, winding branches are visible in the background, their dark\\
silhouettes contrasting against the lighter sky.\\
7:the sky is overcast with grey clouds diffusing the natural light evenly across the landscape.\\
8:the ground is covered with a thick layer of white snow, which appears to be soft and powdery.\\
the surface is uneven with visible footprints and ski tracks.\\} \\

\midrule

        \makecell{\includegraphics[width=6cm]{vis/000000578967.jpg}} &
        \makecell{\includegraphics[width=6cm]{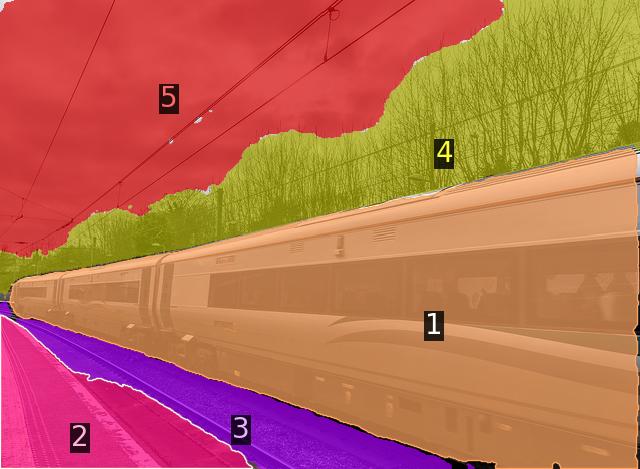}} &
        \makecell[l]{1:stands prominently on the railroad tracks, its sleek silver body contrasting against the blue sky.\\
        its windows reflect the surrounding environment, mirroring the surrounding environment.\\
        the train appears ready for departure, its\\
2:the platform is made of concrete and appears clean and well-maintained. it's where passengers\\
would wait for the train.\\
3:the railroad tracks are metallic and shiny, running horizontally across the image, supporting \\
the train.\\
4:trees with bare branches are visible in the background, their silhouettes contrasting against \\
the sky.\\
5:the sky is overcast, with a greyish hue indicating possibly cloudy weather conditions.\\} \\

\midrule

        \makecell{\includegraphics[width=6cm]{vis/000000134722.jpg}} &
        \makecell{\includegraphics[width=6cm]{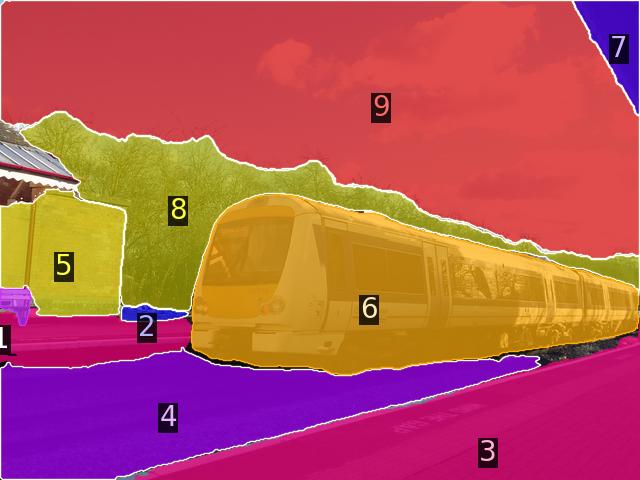}} &
        \makecell[l]{       
1:a bench is visible on the platform ; it's made of wood and metal supports for waiting \\
passengers.\\
2:patches of green grass are visible near the gravel and stone wall, adding a touch of nature to\\
this industrial scene.\\
3:the platform is made of concrete and appears clean and well-maintained. it is where passengers\\
wait for the train.\\
4:the railroad tracks are metallic and shiny, running horizontally across the image, with wooden\\
planks in between the metal rails.\\
5:a brick wall is visible to the right of the image, showcasing an old architectural style with \\
visible wear over time.\\
6:a modern passenger train with a sleek design, painted in white and yellow colors, is captured in\\
motion on the tracks.\\
7:the ceiling is structured with green metal beams supporting it, extending over the platform to\\
provide shelter.\\
8:trees with bare branches are visible in the background, their silhouettes contrasting against the\\
sky.\\
9:the sky is clear with a bright blue hue, indicating a sunny day without any visible clouds.\\
} \\

        \bottomrule
    \end{tabular}
} 

\end{table*}

\begin{table*}[ht]

\begin{minipage}{0.99\textwidth}
\centering  
\scalebox{0.73}{
\begin{tabular}{r p{11.5cm} }
\toprule
& \includegraphics[height=4.5cm]{vip_vis/1.png} \\
\textbf{Question:} & How to cook the dish \(<region 1>\) \\
\textbf{GPT4RoI-7B:} & To cook the dish in the bowl at region1 you need to follow these steps: 1. Heat oil in a large pot or pan. 2. Add onions and cook until softened. 3. Add spices such as curry powder, turmeric, cumin, and coriander. Mix well and cook for a few minutes until fragrant. 4. Add chicken and cook until browned and cooked through. 5. Add vegetables like carrots and green beans. Cook until tender but still crisp. 6. Add the sauce and cook until the chicken is fully cooked and the sauce has thickened. 7. Serve the dish in the bowl at region1 with rice or bread on the side. This dish is a flavorful and hearty meal that can be enjoyed with friends and family. \\
\textbf{GPT4RoI-P2C-7B:} & To cook the dish `chickpea soup', you need to follow these steps: 1. Gather your ingredients: You will need chickpeas, onions, garlic, ginger, tomatoes, and spices. You also have a bowl to serve the soup in and a large pot to cook the soup in. 2. Prepare the chickpeas: Rinse the chickpeas and cook them in a large pot of boiling salted water until they are tender. Drain and set aside. 3. Cook the onions and garlic: Place the onions and garlic in a large pot and cook them until they are soft and translucent. 4. Add the spices: Once the onions and garlic are cooked, add the spices, such as ground turmeric and paprika, to the pot. Stir well to coat the onions and garlic evenly. 5. Add the tomatoes and stir: Drain the chickpeas and add the tomatoes to the pot. Stir well to combine with the onions and garlic. 6. Simmer the soup: Let the soup simmer over low heat for about 30-40 minutes, or until the vegetables are tender and the flavors have melded together. 7. Serve the soup: Serve the soup in a bowl and garnish with fresh herbs, such as cilantro or parsley. 8. Enjoy your delicious chickpea soup!
\\
\textbf{Required capabilities:} & Recognition, Knowledge, Language Generation\\

\midrule

~ &  \includegraphics[height=4.5cm]{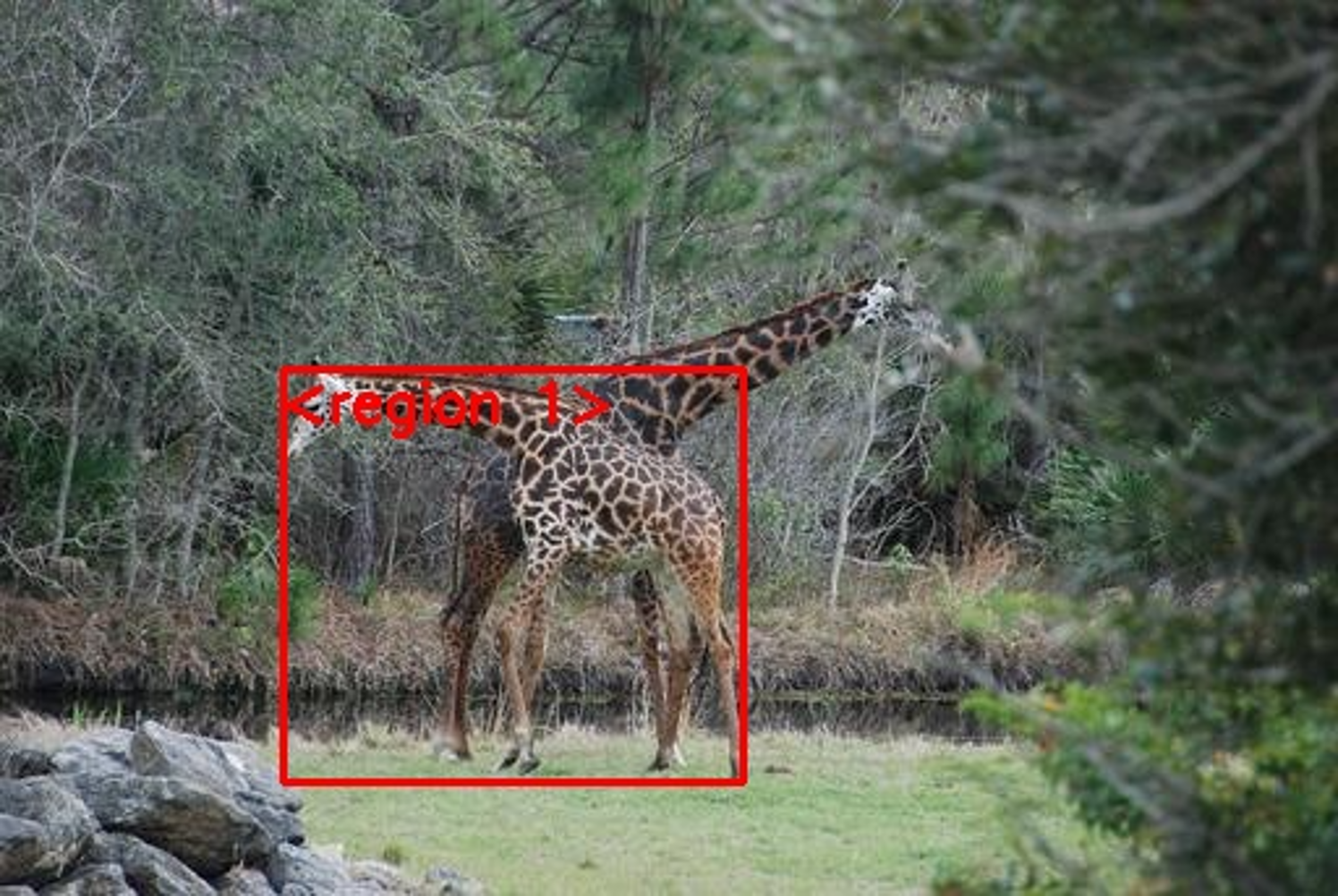} \\

\textbf{Question:} & Is the animal \(<region 1>\) shorter or taller?
  \\
\textbf{GPT4RoI-7B:} & The animal is giraffe at region1. \\
\textbf{GPT4RoI-P2C-7B:} & The animal giraffe at region1 is shorter. \\ 
\textbf{Required capabilities:} & Recognition \\
\midrule

~ & \includegraphics[height=4.5cm]{vip_vis/3.png} \\

\textbf{Question:} & What does the object \(<region 1>\) contain?
  \\
\textbf{GPT4RoI-7B:} & The object contains apple juice. \\
\textbf{GPT4RoI-P2C-7B:} & The object contains peaches. \\ 
\textbf{Required capabilities:} & Recognition \\

\bottomrule
\end{tabular}
}
\end{minipage}
\caption{Additional qualitative comparison on ViP-Bench.  }
\label{tab:vis_vip}  
\end{table*}

\end{document}